\documentclass{article}

\PassOptionsToPackage{square}{natbib}

     \usepackage[preprint]{arxiv}

\usepackage[utf8]{inputenc} 
\usepackage[T1]{fontenc}    
\usepackage{hyperref}
\hypersetup{
    colorlinks=true,
    linkcolor=black,
    citecolor=blue
}
\usepackage{times}
\usepackage{url}
\usepackage{amsfonts}       
\usepackage{nicefrac}       
\usepackage{microtype}      
\usepackage{amsthm}
\usepackage{amsmath,mathtools}
\usepackage{empheq}
\usepackage{amssymb}
\usepackage{booktabs}
\usepackage{multirow}
\usepackage{subcaption}
\usepackage{wrapfig}
\usepackage{chngcntr}
\usepackage{xspace}
\usepackage{bm}
\usepackage[ruled,vlined,linesnumbered]{algorithm2e}
\usepackage{nicefrac,xfrac}
\usepackage{multirow}
\usepackage[table,dvipsnames]{xcolor}
\usepackage{tcolorbox}
\usepackage{makecell}
\usepackage[ruled,vlined,linesnumbered]{algorithm2e}
\usepackage{pifont}

\colorlet{mylinkcolor}{violet}
\colorlet{mycitecolor}{YellowOrange}
\colorlet{myurlcolor}{Aquamarine}

\newcommand{\vx}{\bm{x}}
\newcommand{\vy}{\bm{y}}

\newcommand{\localStep}{\tau}

\newcommand{\activeClients}{\mathcal{S}}
\newcommand{\sgrad}{g}

\newcommand{\localChange}{\Delta}

\newcommand{\lr}{\eta}
\newcommand{\slr}{\alpha}

\usepackage[colorinlistoftodos,prependcaption]{todonotes}

\usepackage{enumitem}
\usepackage{cleveref}
\crefname{equation}{}{}
\Crefname{equation}{}{}
\crefname{thm}{theorem}{theorems}
\Crefname{thm}{Theorem}{Theorems}
\crefname{clm}{claim}{claims}
\Crefname{clm}{Claim}{Claims}
\Crefname{coro}{Corollary}{Corollaries}
\Crefname{lem}{Lemma}{Lemmas}
\Crefname{sec}{Section}{Sections}
\crefname{app}{appendix}{appendices}
\Crefname{app}{Appendix}{Appendices}
\Crefname{part}{Part}{Parts}
\crefname{prop}{proposition}{propositions}
\Crefname{prop}{Proposition}{Propositions}
\Crefname{propty}{Property}{Properties}
\crefname{figure}{fig.}{figures}
\Crefname{figure}{Figure}{Figures}
\crefname{defn}{definition}{definitions}
\Crefname{defn}{Definition}{Definitions}
\crefname{fact}{fact}{facts}
\Crefname{fact}{Fact}{Facts}
\crefname{appendix}{appendix}{appendices}
\Crefname{appendix}{Appendix}{Appendices}
\crefname{algo}{algorithm}{algorithms}
\Crefname{algo}{Algorithm}{Algorithms}
\crefname{algorithm}{algorithm}{algorithms}
\Crefname{algorithm}{Algorithm}{Algorithms}
\crefname{conj}{conjecture}{conjectures}
\Crefname{conj}{Conjecture}{Conjectures}
\crefname{obs}{observation}{observations}
\Crefname{obs}{Observation}{Observations}
\crefname{assump}{assumption}{assumptions}
\Crefname{assump}{Assumption}{Assumptions}
\crefname{rem}{remark}{remarks}
\Crefname{rem}{Remark}{Remarks}

\newcommand{\fedpt}{FedPT }
\renewcommand{\cite}{\citep}
\newcommand{\serveropt}{\text{ServerOpt}\xspace}
\newcommand{\clientopt}{\text{ClientOpt}\xspace}

\title{Efficient and Private Federated Learning \\ with Partially Trainable Networks}

\author{Hakim Sidahmed, Zheng Xu, Ankush Garg, Yuan Cao, Mingqing Chen \\
Google\\
\{hsidahmed, xuzheng, ankugarg, yuancao, mingqing\}@google.com
}

\begin{document}

\maketitle

\begin{abstract}
Federated learning is used for decentralized training of machine learning models on a large number (millions) of edge mobile devices.
It is challenging because mobile devices often have limited communication bandwidth and local computation resources.
Therefore, improving the efficiency of federated learning is critical for scalability and usability.
In this paper, we propose to leverage partially trainable neural networks, which freeze a portion of the model parameters during the entire training process, to reduce the communication cost with little implications on model performance.
Through extensive experiments, we empirically show that Federated learning of Partially Trainable neural networks (FedPT) can result in superior communication-accuracy trade-offs, with up to $46\times$ reduction in communication cost, at a small accuracy cost.
Our approach also enables faster training, with a smaller memory footprint, and better utility for strong differential privacy guarantees.
The proposed \fedpt method can be particularly interesting for pushing the limitations of overparameterization in on-device learning.
\end{abstract}

\section{Introduction}
A large trove of data are being generated with the proliferation of edge devices, such as mobile phones, medical sensors, and smart home devices. 
While those data can be used to develop highly intelligent algorithms, they often contain private information that users are not willing to share with others.
In recent years, Federated Learning (FL) \citep{mcmahan2017communication} has been introduced as an alternative to centralized learning to protect user privacy when training a machine learning model.
In FL, participating clients collaboratively learn a shared model under the supervision of a central server: 
each communication round often starts with the server broadcasting the global model to the participants;
these participants then perform computations on their local data, and send their aggregated updates back to the server to update the global model \citep{kairouz2019advances}.
While FL can be performed on a relatively small number of clients, many applications involve a large number of edge devices, such as mobile phones, or sensors \citep{ramaswamy2020training,sheller2020federated,granqvist2020improving}.
This setting is referred to as cross-device FL.
Training large models on edge devices is challenging due to unreliable connections, and limited computational capabilities.

Federated averaging (FedAvg) \citep{mcmahan2017communication} is one of the most popular algorithms in federated learning.
It was later generalized to a framework of two stage optimization \citep{reddi2020adaptive}:
a client optimizer is used to update local models from the client's local data, 
and a server optimizer is used to update the global model from the aggregated client updates.
Instead of averaging client local models to replace the global model, the client updates (the difference between the initial model received from the server, and the client local model after training on private data) are aggregated, and used as pseudo-gradients to update the global model.
We build our method based on generalized FedAvg, as it can satisfy many restricting constraints of cross-device FL \citep{wang2021field}.

By combining federated learning with differential privacy (DP), a stronger privacy protection can be provided to the clients \citep{kairouz2019advances,kairouz2021practical}.
Specifically, differential privacy can prevent memorization, and protect against potential leakage of user data from a model released publicly \citep{ramaswamy2020training}.
Training large models with DP is known to be hard \citep{tramer2020differentially}, and reducing the dimensionality in DP is an active area of research \citep{zhou2020bypassing,kairouz2021nearly,wang2021field}. 

Deep neural networks have achieved impressive performance on various machine learning tasks, which have become popular in federated learning in recent years \citep{kairouz2019advances,li2020federated,yang2019federated}.
The performance of deep neural networks can often be improved by increasing the model size in the overparameterized regime \citep{allen2019learning,neyshabur2018towards,sankararaman2020impact,du2018gradient}.
Even though parameters of overparameterized networks are believed to be redundant \citep{frankle2019lottery}, and can be pruned, the size of the model is believed to help regularize the optimization landscape to facilitate training.
More recently, researchers have observed that training a small fraction of the parameters of a large model, such as batch normalization layers in convolutional networks \citep{frankle2021training}, can achieve comparable performance as training all the parameters.
Inspired by the recent deep learning research in centralized training, we study the effects of freezing part of the parameters of a large model in federated learning.

In this paper, we propose the use of partially trainable networks (PTNs) to tackle the communication and computation burdens of training large models, which can make FL more accessible to various applications.
Based on the generalized FedAvg algorithm, we only communicate the trainable parameters (which can represent as little as 2\% of the total network parameter count in our study), and a random seed from server to clients.
Clients reconstruct the full model by regenerating the frozen parameters from the random seed, perform local training on private data, and send back the updates on the trainable parameters to the server.
The communication can be significantly saved by the size of frozen parameters. Clients can also save local computations, and memory on gradient calculations for the frozen parameters.
The proposed \fedpt algorithm is used to train various network architectures, including convolutional networks for computer vision, and Transformers for language tasks.
We empirically show the advantages of \fedpt by running experiments on three benchmark datasets: EMNIST, CIFAR-10, and Stack Overflow.
Reducing the communication cost by $1.1\times$ to $46\times$ results in a loss in accuracy between $0.1\%$ and $4\%$.
The simulation training times are reduced by $25\%$, and the memory footprint is reduced by $10\%$ in some settings.
Moreover, differentially private \fedpt achieves better utility (accuracy) than training the full model when the privacy protection is strong (small $\epsilon$).

\section{Related Work}

\paragraph{Communication Efficient Federated Learning}
\citet{mcmahan2017communication} exploit local updates to save the number of communication rounds in FL.
\citet{konevcny2016federated} introduce compression techniques to reduce the uplink communication costs for the clients.
They show a reduction of up to two orders of magnitude in these communication costs.
\citet{caldas2019expanding} apply the lossy compression techniques used in the clients in \citep{konevcny2016federated} to the model on the server.
The authors also introduce federated dropout, in which each client receives a submodel from the server by randomly removing some parameters.
Closely related to federated dropout, ordered dropout \citep{horvath2021fjord} extracts submodels from a main model, and adapts the computation and communication costs to the edge device capabilities.
\citet{singhal2021federated} propose a method to partially reconstruct a model on the edge devices, which results in a reduced communication cost.
The reconstructed parameters are local to the clients, and never sent to the server.
\citet{wang2021pufferfish} proposed training low-rank, pre-factorized deep networks to reduce communication in distributed learning.
Other methods, like compression, and knowledge distillation have been used in FL to reduce the communication costs.
We refer the interested reader to \citep{wang2021field} for more details.
The proposed \fedpt is complementary to these methods, and can be combined with them to further improve the communication efficiency in federated learning.

\paragraph{Partially Trainable Neural Networks}
Partially trainable neural networks have a long history in machine learning.
Extreme learning machines can be considered single layer networks with frozen hidden nodes \citep{huang2015trends}.
Echo state networks \citep{jaeger2002adaptive}, and liquid state machines \citep{maass2002real} disentangle randomly projected inputs with a (generally linear) readout layer. 
More recently, \citet{zhang2019layers} studied the properties of layers in overparameterized deep models.
They distinguish \emph{ambient} layers from \emph{critical} ones, and show that the ambient layers are robust to re-randomization and re-initialization in the post-training validation.
\citet{rosenfeld2018intriguing} show competitive results on vision tasks where they freeze most of the weights of the model.
\citet{frankle2021training} only train the parameters of batch normalization layers of vision models, and freeze the other weights.
Randomized networks have also been studied for language tasks, such as sentence classification \citep{wietingrandom2019}, machine translation \citep{garg2020echo, shen2021reservoir}, and speech recognition \citep{harshechospeech}, where learning only linear mappings on top of random features achieves surprisingly high accuracy.
We are not aware of any previous work applying PTNs to federated learning.

\paragraph{Efficient Transformers}
A lot of attention has been put lately on designing efficient architectures for Transformers \citep{vaswani2017attention, fournier2021practical}.
FNets \citep{leethorp2021fnet} replace the self-attention layer by a Fourier transform, reducing its computational complexity from quadratic in the input sequence size to linear. 
\citet{shen2021reservoir} use random layers as a computationally efficient way to increase network depth.
They intersperse reservoir frozen Transformer layers with regular, trainable ones.
They also prove that backward pass can be skipped altogether for these layers.
\citet{choromanski2021rethinking} approximate the attention using a kernel with randomized mappings.
\citet{wang2020linformer} approximate the attention with a low-rank factorization.
\citet{child2019generating} introduce the Sparse Transformer, in which the attention matrix is approximated with a sparse factorization.
We apply \fedpt to efficiently train Transformers in federated learning.

\paragraph{Model Compression}
Pruning \citep{lecun1990optimal} reduces the model size by deleting the weights whose removal have a minimal effect on the model performance.
It was observed in \citet{frankle2019lottery} that sub-networks can be found in a dense neural network that can match the test accuracy of the original network.
Knowledge distillation \citep{ba2014deep, hinton2015distilling} transfers the knowledge of a large teacher into a smaller student, which can then be used at inference time.
While the methods mentioned can be leveraged to make the inference of federated trained models lighter, they do not alleviate the bottlenecks present during on-device training.
Our proposed FedPT, on the other hand, aims at making the training process more efficient.

\section{Partially Trainable Networks for Federated Learning}

\subsection{\fedpt Method}

\begin{algorithm}[ht]
    \DontPrintSemicolon
    \SetKwInput{Input}{Input}
    \SetAlgoLined
    \LinesNumbered
    \Input{Initial model $\vx^{(0)}$; \clientopt, \serveropt with learning rates $\lr, \slr$}
    Split $\vx^{(0)}$ into trainable part $\vy^{(0)}$, and non-trainable part generated by random seed $z$
    
     \For{$t \in \{0,1,\dots,T-1\}$ }{
      
      Send  $(\vy^{(t)}, z)$ to a subset $\activeClients^{(t)}$ of clients
      
      \For{{\it \bf client} $i \in \activeClients^{(t)}$ {\it \bf in parallel}}{
        Initialize local model $\vx_i^{(t,0)}=\text{Reconstruct} (\vy^{(t)}, z)$\;
        \For {$k =0,\dots,\localStep_i-1$}{
            Compute local stochastic gradient $\sgrad_i(\vy_i^{(t,k)})$ by backprop through $\vx_i^{(t,k)}$\;
            Perform local update $\vy_i^{(t,k+1)} = \clientopt(\vy_i^{(t,k)}, \sgrad_i(\vy_i^{(t,k)}), \lr, t)$\;
        }
        Compute and send back local model changes $\localChange_i^{(t)} = \vy_i^{(t,\localStep_i)} - \vy_i^{(t,0)}$\;
      }
      Aggregate local changes $\localChange^{(t)} = \sum_{i \in \activeClients^{(t)}} p_i \localChange_i^{(t)} / \sum_{i \in \activeClients^{(t)}} p_i$\;
      Update global model $\vy^{(t+1)} = \serveropt(\vy^{(t)}, -\localChange^{(t)},\slr,t)$\;
     }
     \caption{FedPT: federated learning of partially trainable neural networks}
     \label{algo:fedpt}
\end{algorithm}

We use PTNs in federated learning tasks in the proposed \fedpt algorithm. 
Inspired by previous partially trainable models in the centralized setting \citep{huang2015trends, Giryes_2016}, we freeze a set of parameters after initialization.
This allows us to summarize the frozen parameters into a single random seed, provided that the server and clients share the same random number generator.
This seed is sent to the clients (edge devices), and used to reconstruct the frozen parameters. Clients do not need to send updates for the frozen parameters back.
We summarized the method in \cref{algo:fedpt}, which is based on the generalized FedAvg algorithm with two stage optimization by \serveropt and \clientopt.
In cross-device FL, only a small subset of the clients $\activeClients^{(t)}$ (compared to the large population) can be accessed at each communication round $t$.
We use the number of local samples on client $i$ as weight $p_i$ to aggregate the local updates following \citep{mcmahan2017communication}.  

The design of PTNs highly depends on the network architecture.
Since freezing a small number of parameters does not impact the communication efficiency notably, we focus on freezing layers that contain a large proportion of the parameters in our experiments, and choose different layers for different architectures.
To maximize the communication and computation efficiencies of the PTN, we suggest the following principles for experimental design:
\begin{enumerate}
    \item Start from freezing the largest parameter block of a network.
    \item Add more blocks if it did not degrade the model performance on utility (accuracy).
    \item Switch to a smaller block if it did degrade the model performance by a large margin.
    \item Repeat these steps to find the best PTN.
\end{enumerate}
Once an optimal PTN is found for a specific network architecture, the same PTN can be used for various application tasks that use the same network architecture.

We study several popular network architectures:
ResNet \citep{he2016deep} with group normalization for image tasks, small convolutional neural networks as feature extractors with a few fully connected layers for classification, and Transformer \citep{vaswani2017attention} for language tasks.
We experiment with freezing the following layers, and test on FL benchmark datasets:
\begin{itemize}
    \item Convolutional layers in resnet-18 trained on CIFAR-10.
    \item Encoder dense layers in the Transformer architecture used in Stack Overflow.
    \item Dense layer following the convolutional layers in the convolutional model used in EMNIST.
\end{itemize}

The networks and benchmarks are introduced in \citep{mcmahan2017communication, hsu2019measuring, reddi2020adaptive, wang2021field}, and are widely used for cross-device FL. 
We notice that freezing the normalization layers in \fedpt notably hurts the performance of ResNet, as observed in \citep{frankle2021training} in the centralized setting.
Since the normalization layers usually have a small number of parameters, we always train them in FedPT.
The frozen parameters are set to their initial values, which are generated from Gaussian initializers.

\subsection{Advantages of \fedpt}
\paragraph{Communication Efficiency}
Communication is one of the main bottlenecks in cross-device federated learning.
Model transmission from server to devices can be a serious constraint for the server,  particularly when stragglers that have limited network connections exist.
Sending the model updates back to the server can be even more challenging, as uplink is typically much slower than downlink.
The download and upload bandwidths in a real cross-device FL system are estimated at $0.75$MB/s and $0.25$MB/s respectively by \citet{wang2021field}. 
\fedpt can mitigate this issue: the frozen parameters are compressed into a random seed sent from server to clients, and participating clients do not need to send any updates back for these frozen parameters.

\paragraph{Differential Privacy (DP)}
Federated learning is designed for privacy protection, as the clients do not share their private data.
Combining FL with DP can provide even stronger privacy defenses \citep{kairouz2019advances,wang2021field}.
The most popular method to achieve user-level DP in FL is a variant of generalized FedAvg, called DP-FedAvg, which clips the model updates $\Delta_i^{(t)}$, and adds noise to the aggregated $\Delta^{(t)}$ \citep{mcmahan2017learning}.
It is nontrivial to get good utility-privacy trade-offs when applying DP in FL, as it is challenging for practical cross-device FL to perform sampling for privacy amplification \citep{wang2021field}.
Recently, \citet{kairouz2021practical} proposed to use the Differentially Private Follow the Regularized Leader (DP-FTRL) method on the server in the generalized FedAvg framework to get formal $(\epsilon, \delta)$-DP guarantees.
We combine DP-FTRL with \fedpt to get $(\epsilon, \delta)$ privacy guarantees.
In FedPT, we only need to make the trainable parameters private, and this dimension reduction effect can be beneficial for differential privacy.  
We only clip the client updates $\Delta_i^{(t)}$, and add noise to the aggregated $\Delta^{(t)}$ of trainable parameters, which can potentially tolerate smaller clip norms, and larger noises. 
In our experiments, partially trained models are more resilient to high noise levels than fully trained ones.

\paragraph{Training time}
Reducing the client training time allows more devices to complete their local computations in the allotted time in a round, which is often desirable in practical FL systems \citep{bonawitz2019towards,wang2021field}.
In addition to that, reducing the training time makes it possible to train larger models in production settings where the federated learning tasks have a limited amount of time to run on edge devices.
\fedpt can reduce the training time, as it does not need to calculate gradients for the frozen parameters. 
We observe, in some of our simulations, that \fedpt has a reduction in the client training time as the number of frozen parameters increases.
Our study shows that \fedpt has a significant decrease in runtime for deep convolutional models, for which we freeze the convolutional layers.

\paragraph{Memory footprint}
\fedpt can also improve the memory footprint of model training in FL, as we do not need to save intermediate activations of frozen layers after backprop, and we do not need to calculate or save the gradients for the frozen layers.
Besides, when computing the model updates on clients, two copies of the trainable parameters (new value, and old one) are needed to generate the client model update $\Delta_i$, but none  is needed for the frozen ones.
We compare the memory footprints of fully/partially trainable models, and see a reduction in the settings where we freeze the convolutional layers.

\section{Experiments}

\paragraph{Setup}
We use three datasets: EMNIST \footnote{\url{https://www.tensorflow.org/federated/api_docs/python/tff/simulation/datasets/emnist}} \citep{cohen2017emnist}, CIFAR-10 \citep{krizhevsky2009learning}, and Stack Overflow \footnote{\url{https://www.tensorflow.org/federated/api_docs/python/tff/simulation/datasets/stackoverflow}}.
We follow \citep{hsu2019measuring} to use a Dirichlet distribution of parameter 1 to create a heterogeneous dataset for CIFAR-10, and \citep{wang2021field} to preprocess this dataset.
We follow \citep{reddi2020adaptive,wang2021field} to preprocess the EMNIST, and Stack Overflow datasets.
EMNIST and CIFAR-10 are image datasets, for which we train convolutional networks.
Stack Overflow is a language dataset, for which we train a Transformer-based next word prediction model. We measure runtime and memory footprint on desktop CPUs to simulate on-device training. More details on models and hyperparameters used are provided in the Appendix.


\subsection{Efficiency of \fedpt}

\begin{table}[th]
  \caption{Percentage of trainable parameters, reduction in communication cost, average accuracies, and training times for EMNIST.}
  \label{emnist-results}
  \centering
  \begin{tabular}{llll}
    \toprule
    Trainable Parameters (\%)              &     Reduction in Communication         &    Final Accuracy      & Runtime (s)        \\
    \midrule
    $4.97$                                &         $20\times$                   &   $83.75 \pm 0.16$     &  $2.90 \pm 0.42$   \\
    $100$                                 &         $1\times$                    &   $85.47 \pm 0.21$     &  $2.72 \pm 0.37$   \\
    \bottomrule
  \end{tabular}
\end{table}

\begin{table}[th]
  \caption{Percentage of trainable parameters, reduction in communication cost, average accuracies, and training times for CIFAR-10.}
  \label{cifar-10-results}
  \centering
  \begin{tabular}{llll}
    \toprule
    Trainable Parameters (\%)    & Reduction in Communication    & Final Accuracy        & Runtime (s)           \\
    \midrule
    $2.16$                      & $46.3\times$                  &   $79.49 \pm 1.1$     &  $ 16.22 \pm 0.076 $  \\
    $3.47$                      & $28.8\times$                  &   $80.76 \pm 0.72$    &  $ 16.58 \pm 0.053 $  \\
    $8.07$                      & $12.4\times$                  &   $82.82 \pm 0.18$    &  $ 16.96 \pm 0.021 $  \\
    $26.25$                     & $3.8\times$                   &   $83.57 \pm 0.41$    &  $ 17.89 \pm 0.057 $  \\
    $100$                       & $1.0\times$                   &   $83.65 \pm 0.26$    &  $ 20.51 \pm 0.080 $  \\
    \bottomrule
  \end{tabular}
\end{table}

\paragraph{EMNIST with CNN} Table \ref{emnist-results} shows the results for EMNIST.
We freeze the dense layer following the convolutional ones for this model presented in table \ref{emnist-model}.
The reduction in communication cost is 20 fold, for an absolute loss in accuracy of less than 2\%, and similar runtimes.

\paragraph{CIFAR-10 with ResNet} Table \ref{cifar-10-results} summarizes the results for the CIFAR-10 task, trained on resnet-18.
The training time per round monotonically decreases as we freeze more parameters.
While the accuracy decreases as well, a reduction of $4 \times$ in the communication costs results in almost the same accuracy.

\begin{table}[th]
  \caption{Percentage of trainable parameters, reduction in communication cost, average accuracies, and training times for Stack Overflow.}
  \label{sonwp-results}
  \centering
  \begin{tabular}{llll}
    \toprule
    Trainable Parameters (\%)                & Reduction in Communication          & Final Accuracy           & Runtime (s)           \\
    \midrule
    $73.8$                                  & $1.4\times$                       &   $23.23 \pm 0.22$       & $9.37 \pm 0.34$        \\
    $82.6$                                  & $1.2\times$                       &   $23.71 \pm 0.16$       & $9.23 \pm 0.46$        \\
    $91.3$                                  & $1.1\times$                       &   $24.09 \pm 0.16$       & $9.42 \pm 0.55$        \\
    $100$                                   & $1.0\times$                       &   $24.50 \pm 0.19$       & $9.74 \pm 0.46$        \\
    \bottomrule
  \end{tabular}
\end{table}

\paragraph{Stack Overflow with Transformer} Table \ref{sonwp-results} shows the results for Stack Overflow Next Word Prediction.
In this task, the training times are not sensibly changed, and we can trade-off 10\% of communication costs for less than $0.5\%$ accuracy.

\paragraph{Memory footprint}
In order to verify our intuition that the memory footprint can be positively impacted by FedPT, we profile the memory in the CIFAR-10 task in simulation experiments.
Our results, summarized in table \ref{cifar-10-peak-memories}, show a reduction in the peak memory usage as the proportion of frozen parameters increases.

\begin{table}[th]
  \caption{Peak memory at different percentages of trainable parameters for CIFAR-10.}
  \label{cifar-10-peak-memories}
  \centering
  \begin{tabular}{ll}
    \toprule
    Trainable Parameters (\%)                & Peak Memory (MiB)           \\
    \midrule
    $2.16$                                  & $505.94$               \\
    $3.47$                                  & $506.37$              \\
    $8.07$                                  & $506.37$              \\
    $26.25$                                  & $520.43$              \\
    $100$                                   & $565.43$              \\
    \bottomrule
  \end{tabular}
\end{table}

\subsection{Differential Privacy}
We experiment with DP-FTRL \citep{kairouz2021practical} on the Stack Overflow next word prediction task.
We use the same noise multipliers, number of clients per round (100), and total number of rounds (1600) as in \citep{kairouz2021practical}, and hence achieve the same $(\epsilon, \delta)$ privacy guarantees.
The results are presented in \Cref{dpftrl-1-results}.
We observe that FedPT is more resilient to high levels of noise, making it a better fit to offer strong privacy guarantees.
An intuitive explanation is that, by updating less parameters, \fedpt is less affected by a given noise level (the noise is added to less parameters).



\begin{table}[th]
  \caption{Training accuracies (\%) for Fully Trainable (FT), and Partially Trainable (PT) models with DP.}
  \label{dpftrl-1-results}
  \centering
  \begin{tabular}{lllllll}
    \toprule
    Noise multiplier     & $0.0$     & $1.13$    & $2.33$    & $4.03$    & $6.21$    & $8.83$ \\
    Privacy $\epsilon$ & $\infty$ & 18.71 & 7.83 & 4.19 & 2.60 & 1.77\\ 
    \midrule
    FT   & 23.22                  & 19.67            & 18.37            & 17.28            & 16.24             & 14.60 \\
    PT     & 22.44                & 19.58            & 18.38            & 17.03            & \textbf{16.28}            & \textbf{15.01} \\
    \bottomrule
  \end{tabular}
\end{table}

\section{Discussion and Future Work}
Our experiments show that PTNs can be used to improve the efficiency and privacy in FL.
By leveraging the ability to summarize the frozen parameters into a random seed when broadcasting them from the server to the clients, the communication cost can be drastically reduced.
Reducing the number of trainable parameters also has implications for the privacy guarantees.
We suspect that the accuracy gap we observe in our experiments between fully trainable, and partially trainable models can be reduced in multiple ways, such as running a grid search to find the optimal parameters for the partially trainable parameters, or adding layers of frozen parameters.

We present extensive simulation experiments to suggest PTNs can have a positive impact on real world FL applications. Note that our profiling experiments ran in a data center for simulation, while real-world training happens on edge devices for cross-device FL. We are interested in studying the effects on memory and runtime in real-world on-device training, which may depend on the network architectures, the parameters which are frozen, and implementation details.
Though the on-device performance can be reasonably inferred from data center simulation \citep{wang2021field}, the data center and on-device settings can use different computation optimizations in implementation and hardware.

We are also interested in combining FedPT with other methods for improving communication and computation efficiency, such as federated dropout \citep{caldas2019expanding}, to further ease the communication bottleneck inherent to FL.
Furthermore, we presume that \fedpt can benefit from better generalization bounds, similar to the ones shown in \citep{zhang2019layers} based on parameter counting \citep{10.5555/1795646}.
The application of \fedpt to edge devices can result in more clients contributing to the training of the global model, potentially leading to better fairness.

Another possible extension to our work is to selectively freeze more parameters for devices with smaller bandwidth and/or computational capacity, while training more parameters on devices that do not suffer such limitations.
Finally, a systematic study of the relationship between the final test accuracy, and the architecture-specific frozen layers can provide more insights and guidelines for improving FedPT's performance.

\section{Conclusion}
This paper introduces \fedpt for training partially trainable neural networks in the context of federated learning.
Our empirical results suggest that it is possible to achieve competitive accuracy using federated models when a significant portion of the parameters are untrained, which improves efficiency in terms of communication cost, and computational complexity.
We also show that partially trainable models can be more resilient to high additive noise, enabling better utility for stronger privacy guarantees than fully trained models.
Based on these findings, \fedpt is a promising method to improve efficiency and privacy for practical federated learning applications.

\section*{Acknowledgements}
The authors would like to thank Zachary Garrett, Jakub Konečný, Keith Rush, Changwan Ryu, Om Thakkar, and  Petr Zadrazil for helpful discussions.

\bibliographystyle{plainnat}
{\bibliography{references.bib}}

\begin{thebibliography}{53}
\providecommand{\natexlab}[1]{#1}
\providecommand{\url}[1]{\texttt{#1}}
\expandafter\ifx\csname urlstyle\endcsname\relax
  \providecommand{\doi}[1]{doi: #1}\else
  \providecommand{\doi}{doi: \begingroup \urlstyle{rm}\Url}\fi

\bibitem[Allen-Zhu et~al.(2019)Allen-Zhu, Li, and Liang]{allen2019learning}
Zeyuan Allen-Zhu, Yuanzhi Li, and Yingyu Liang.
\newblock Learning and generalization in overparameterized neural networks,
  going beyond two layers.
\newblock \emph{Advances in neural information processing systems}, 2019.

\bibitem[Anthony and Bartlett(2009)]{10.5555/1795646}
Martin Anthony and Peter~L. Bartlett.
\newblock \emph{Neural Network Learning: Theoretical Foundations}.
\newblock Cambridge University Press, USA, 1st edition, 2009.
\newblock ISBN 052111862X.

\bibitem[Ba and Caruana(2014)]{ba2014deep}
Lei~Jimmy Ba and Rich Caruana.
\newblock Do deep nets really need to be deep?, 2014.

\bibitem[Bonawitz et~al.(2019)Bonawitz, Eichner, Grieskamp, Huba, Ingerman,
  Ivanov, Kiddon, Kone{\v{c}}n{\`y}, Mazzocchi, McMahan,
  et~al.]{bonawitz2019towards}
Keith Bonawitz, Hubert Eichner, Wolfgang Grieskamp, Dzmitry Huba, Alex
  Ingerman, Vladimir Ivanov, Chloe Kiddon, Jakub Kone{\v{c}}n{\`y}, Stefano
  Mazzocchi, H~Brendan McMahan, et~al.
\newblock Towards federated learning at scale: System design.
\newblock \emph{SysML}, 2019.

\bibitem[Caldas et~al.(2019{\natexlab{a}})Caldas, Duddu, Wu, Li, Konečný,
  McMahan, Smith, and Talwalkar]{caldas2019leaf}
Sebastian Caldas, Sai Meher~Karthik Duddu, Peter Wu, Tian Li, Jakub Konečný,
  H.~Brendan McMahan, Virginia Smith, and Ameet Talwalkar.
\newblock Leaf: A benchmark for federated settings, 2019{\natexlab{a}}.

\bibitem[Caldas et~al.(2019{\natexlab{b}})Caldas, Konečny, McMahan, and
  Talwalkar]{caldas2019expanding}
Sebastian Caldas, Jakub Konečny, H.~Brendan McMahan, and Ameet Talwalkar.
\newblock Expanding the reach of federated learning by reducing client resource
  requirements, 2019{\natexlab{b}}.

\bibitem[Child et~al.(2019)Child, Gray, Radford, and
  Sutskever]{child2019generating}
Rewon Child, Scott Gray, Alec Radford, and Ilya Sutskever.
\newblock Generating long sequences with sparse transformers, 2019.

\bibitem[Choromanski et~al.(2021)Choromanski, Likhosherstov, Dohan, Song, Gane,
  Sarlos, Hawkins, Davis, Mohiuddin, Kaiser, Belanger, Colwell, and
  Weller]{choromanski2021rethinking}
Krzysztof Choromanski, Valerii Likhosherstov, David Dohan, Xingyou Song,
  Andreea Gane, Tamas Sarlos, Peter Hawkins, Jared Davis, Afroz Mohiuddin,
  Lukasz Kaiser, David Belanger, Lucy Colwell, and Adrian Weller.
\newblock Rethinking attention with performers, 2021.

\bibitem[Cohen et~al.(2017)Cohen, Afshar, Tapson, and van
  Schaik]{cohen2017emnist}
Gregory Cohen, Saeed Afshar, Jonathan Tapson, and André van Schaik.
\newblock Emnist: an extension of mnist to handwritten letters, 2017.

\bibitem[Du et~al.(2018)Du, Zhai, Poczos, and Singh]{du2018gradient}
Simon~S Du, Xiyu Zhai, Barnabas Poczos, and Aarti Singh.
\newblock Gradient descent provably optimizes over-parameterized neural
  networks.
\newblock \emph{arXiv preprint arXiv:1810.02054}, 2018.

\bibitem[Fournier et~al.(2021)Fournier, Caron, and
  Aloise]{fournier2021practical}
Quentin Fournier, Gaétan~Marceau Caron, and Daniel Aloise.
\newblock A practical survey on faster and lighter transformers, 2021.

\bibitem[Frankle and Carbin(2019)]{frankle2019lottery}
Jonathan Frankle and Michael Carbin.
\newblock The lottery ticket hypothesis: Finding sparse, trainable neural
  networks, 2019.

\bibitem[Frankle et~al.(2021)Frankle, Schwab, and Morcos]{frankle2021training}
Jonathan Frankle, David~J. Schwab, and Ari~S. Morcos.
\newblock Training batchnorm and only batchnorm: On the expressive power of
  random features in cnns, 2021.

\bibitem[Garg et~al.(2020)Garg, Cao, and Ge]{garg2020echo}
Ankush Garg, Yuan Cao, and Qi~Ge.
\newblock Echo state neural machine translation, 2020.

\bibitem[Giryes et~al.(2016)Giryes, Sapiro, and Bronstein]{Giryes_2016}
Raja Giryes, Guillermo Sapiro, and Alex~M. Bronstein.
\newblock Deep neural networks with random gaussian weights: A universal
  classification strategy?
\newblock \emph{IEEE Transactions on Signal Processing}, 64\penalty0
  (13):\penalty0 3444–3457, Jul 2016.
\newblock ISSN 1941-0476.
\newblock \doi{10.1109/tsp.2016.2546221}.
\newblock URL \url{http://dx.doi.org/10.1109/TSP.2016.2546221}.

\bibitem[Granqvist et~al.(2020)Granqvist, Seigel, van Dalen, Áine Cahill,
  Shum, and Paulik]{granqvist2020improving}
Filip Granqvist, Matt Seigel, Rogier van Dalen, Áine Cahill, Stephen Shum, and
  Matthias Paulik.
\newblock Improving on-device speaker verification using federated learning
  with privacy, 2020.

\bibitem[He et~al.(2016)He, Zhang, Ren, and Sun]{he2016deep}
Kaiming He, Xiangyu Zhang, Shaoqing Ren, and Jian Sun.
\newblock Deep residual learning for image recognition.
\newblock In \emph{Proceedings of the IEEE conference on computer vision and
  pattern recognition}, pages 770--778, 2016.

\bibitem[Hinton et~al.(2015)Hinton, Vinyals, and Dean]{hinton2015distilling}
Geoffrey Hinton, Oriol Vinyals, and Jeff Dean.
\newblock Distilling the knowledge in a neural network, 2015.

\bibitem[Horvath et~al.(2021)Horvath, Laskaridis, Almeida, Leontiadis,
  Venieris, and Lane]{horvath2021fjord}
Samuel Horvath, Stefanos Laskaridis, Mario Almeida, Ilias Leontiadis,
  Stylianos~I Venieris, and Nicholas~D Lane.
\newblock Fjord: Fair and accurate federated learning under heterogeneous
  targets with ordered dropout.
\newblock \emph{arXiv preprint arXiv:2102.13451}, 2021.

\bibitem[Hsieh et~al.(2020)Hsieh, Phanishayee, Mutlu, and
  Gibbons]{hsieh2020noniid}
Kevin Hsieh, Amar Phanishayee, Onur Mutlu, and Phillip~B. Gibbons.
\newblock The non-iid data quagmire of decentralized machine learning, 2020.

\bibitem[Hsu et~al.(2019)Hsu, Qi, and Brown]{hsu2019measuring}
Tzu-Ming~Harry Hsu, Hang Qi, and Matthew Brown.
\newblock Measuring the effects of non-identical data distribution for
  federated visual classification, 2019.

\bibitem[Huang et~al.(2015)Huang, Huang, Song, and You]{huang2015trends}
Gao Huang, Guang-Bin Huang, Shiji Song, and Keyou You.
\newblock Trends in extreme learning machines: A review.
\newblock \emph{Neural Networks}, 61:\penalty0 32--48, 2015.

\bibitem[Ingerman and Ostrowski(2019)]{tff2019}
Alex Ingerman and Krzys Ostrowski.
\newblock Tensorflow federated.
\newblock
  \url{https://medium.com/tensorflow/introducing-tensorflow-federated-a4147aa20041},
  2019.

\bibitem[Jaeger(2002)]{jaeger2002adaptive}
Herbert Jaeger.
\newblock Adaptive nonlinear system identification with echo state networks.
\newblock \emph{Advances in neural information processing systems},
  15:\penalty0 609--616, 2002.

\bibitem[Kairouz et~al.(2019)Kairouz, McMahan, Avent, Bellet, Bennis, Bhagoji,
  Bonawitz, Charles, Cormode, Cummings, et~al.]{kairouz2019advances}
Peter Kairouz, H~Brendan McMahan, Brendan Avent, Aur{\'e}lien Bellet, Mehdi
  Bennis, Arjun~Nitin Bhagoji, Kallista Bonawitz, Zachary Charles, Graham
  Cormode, Rachel Cummings, et~al.
\newblock Advances and open problems in federated learning.
\newblock \emph{arXiv preprint arXiv:1912.04977}, 2019.

\bibitem[Kairouz et~al.(2021{\natexlab{a}})Kairouz, Diaz, Rush, and
  Thakurta]{kairouz2021nearly}
Peter Kairouz, Monica~Ribero Diaz, Keith Rush, and Abhradeep Thakurta.
\newblock (nearly) dimension independent private erm with adagrad rates via
  publicly estimated subspaces.
\newblock In \emph{Conference on Learning Theory}, pages 2717--2746. PMLR,
  2021{\natexlab{a}}.

\bibitem[Kairouz et~al.(2021{\natexlab{b}})Kairouz, McMahan, Song, Thakkar,
  Thakurta, and Xu]{kairouz2021practical}
Peter Kairouz, Brendan McMahan, Shuang Song, Om~Thakkar, Abhradeep Thakurta,
  and Zheng Xu.
\newblock Practical and private (deep) learning without sampling or shuffling,
  2021{\natexlab{b}}.

\bibitem[Kone{\v{c}}n{\`y} et~al.(2016)Kone{\v{c}}n{\`y}, McMahan, Yu,
  Richt{\'a}rik, Suresh, and Bacon]{konevcny2016federated}
Jakub Kone{\v{c}}n{\`y}, H~Brendan McMahan, Felix~X Yu, Peter Richt{\'a}rik,
  Ananda~Theertha Suresh, and Dave Bacon.
\newblock Federated learning: Strategies for improving communication
  efficiency.
\newblock \emph{arXiv preprint arXiv:1610.05492}, 2016.

\bibitem[Krizhevsky et~al.(2009)Krizhevsky, Hinton,
  et~al.]{krizhevsky2009learning}
Alex Krizhevsky, Geoffrey Hinton, et~al.
\newblock Learning multiple layers of features from tiny images.
\newblock 2009.

\bibitem[LeCun et~al.(1990)LeCun, Denker, and Solla]{lecun1990optimal}
Yann LeCun, John~S Denker, and Sara~A Solla.
\newblock Optimal brain damage.
\newblock In \emph{Advances in neural information processing systems}, pages
  598--605, 1990.

\bibitem[Lee-Thorp et~al.(2021)Lee-Thorp, Ainslie, Eckstein, and
  Ontanon]{leethorp2021fnet}
James Lee-Thorp, Joshua Ainslie, Ilya Eckstein, and Santiago Ontanon.
\newblock Fnet: Mixing tokens with fourier transforms, 2021.

\bibitem[Li et~al.(2020)Li, Sahu, Talwalkar, and Smith]{li2020federated}
Tian Li, Anit~Kumar Sahu, Ameet Talwalkar, and Virginia Smith.
\newblock Federated learning: Challenges, methods, and future directions.
\newblock \emph{IEEE Signal Processing Magazine}, 37\penalty0 (3):\penalty0
  50--60, 2020.

\bibitem[Maass et~al.(2002)Maass, Natschl{\"a}ger, and Markram]{maass2002real}
Wolfgang Maass, Thomas Natschl{\"a}ger, and Henry Markram.
\newblock Real-time computing without stable states: A new framework for neural
  computation based on perturbations.
\newblock \emph{Neural computation}, 14\penalty0 (11):\penalty0 2531--2560,
  2002.

\bibitem[McMahan et~al.(2017{\natexlab{a}})McMahan, Moore, Ramage, Hampson, and
  y~Arcas]{mcmahan2017communication}
Brendan McMahan, Eider Moore, Daniel Ramage, Seth Hampson, and Blaise~Aguera
  y~Arcas.
\newblock Communication-efficient learning of deep networks from decentralized
  data.
\newblock In \emph{Artificial intelligence and statistics}, pages 1273--1282.
  PMLR, 2017{\natexlab{a}}.

\bibitem[McMahan et~al.(2017{\natexlab{b}})McMahan, Ramage, Talwar, and
  Zhang]{mcmahan2017learning}
H~Brendan McMahan, Daniel Ramage, Kunal Talwar, and Li~Zhang.
\newblock Learning differentially private recurrent language models.
\newblock \emph{arXiv preprint arXiv:1710.06963}, 2017{\natexlab{b}}.

\bibitem[Neyshabur et~al.(2018)Neyshabur, Li, Bhojanapalli, LeCun, and
  Srebro]{neyshabur2018towards}
Behnam Neyshabur, Zhiyuan Li, Srinadh Bhojanapalli, Yann LeCun, and Nathan
  Srebro.
\newblock Towards understanding the role of over-parametrization in
  generalization of neural networks.
\newblock \emph{arXiv preprint arXiv:1805.12076}, 2018.

\bibitem[Ramaswamy et~al.(2020)Ramaswamy, Thakkar, Mathews, Andrew, McMahan,
  and Beaufays]{ramaswamy2020training}
Swaroop Ramaswamy, Om~Thakkar, Rajiv Mathews, Galen Andrew, H~Brendan McMahan,
  and Fran{\c{c}}oise Beaufays.
\newblock Training production language models without memorizing user data.
\newblock \emph{arXiv preprint arXiv:2009.10031}, 2020.

\bibitem[Reddi et~al.(2020)Reddi, Charles, Zaheer, Garrett, Rush,
  Kone{\v{c}}n{\`y}, Kumar, and McMahan]{reddi2020adaptive}
Sashank Reddi, Zachary Charles, Manzil Zaheer, Zachary Garrett, Keith Rush,
  Jakub Kone{\v{c}}n{\`y}, Sanjiv Kumar, and H~Brendan McMahan.
\newblock Adaptive federated optimization.
\newblock \emph{arXiv preprint arXiv:2003.00295}, 2020.

\bibitem[Rosenfeld and Tsotsos(2018)]{rosenfeld2018intriguing}
Amir Rosenfeld and John~K. Tsotsos.
\newblock Intriguing properties of randomly weighted networks: Generalizing
  while learning next to nothing, 2018.

\bibitem[Sankararaman et~al.(2020)Sankararaman, De, Xu, Huang, and
  Goldstein]{sankararaman2020impact}
Karthik~Abinav Sankararaman, Soham De, Zheng Xu, W~Ronny Huang, and Tom
  Goldstein.
\newblock The impact of neural network overparameterization on gradient
  confusion and stochastic gradient descent.
\newblock In \emph{International Conference on Machine Learning}, pages
  8469--8479. PMLR, 2020.

\bibitem[Sheller et~al.(2020)Sheller, Edwards, Reina, Martin, Pati, Kotrotsou,
  Milchenko, Xu, Marcus, Colen, et~al.]{sheller2020federated}
Micah~J Sheller, Brandon Edwards, G~Anthony Reina, Jason Martin, Sarthak Pati,
  Aikaterini Kotrotsou, Mikhail Milchenko, Weilin Xu, Daniel Marcus, Rivka~R
  Colen, et~al.
\newblock Federated learning in medicine: facilitating multi-institutional
  collaborations without sharing patient data.
\newblock \emph{Scientific reports}, 10\penalty0 (1):\penalty0 1--12, 2020.

\bibitem[Shen et~al.(2021)Shen, Baevski, Morcos, Keutzer, Auli, and
  Kiela]{shen2021reservoir}
Sheng Shen, Alexei Baevski, Ari~S. Morcos, Kurt Keutzer, Michael Auli, and
  Douwe Kiela.
\newblock Reservoir transformers, 2021.

\bibitem[Shrivastava et~al.(2021)Shrivastava, Garg, Cao, Zhang, and
  Sainath]{harshechospeech}
Harsh Shrivastava, Ankush Garg, Yuan Cao, Yu~Zhang, and Tara Sainath.
\newblock Echo state speech recognition.
\newblock In \emph{ICASSP 2021 - 2021 IEEE International Conference on
  Acoustics, Speech and Signal Processing (ICASSP)}, pages 5669--5673, 2021.
\newblock \doi{10.1109/ICASSP39728.2021.9414495}.

\bibitem[Singhal et~al.(2021)Singhal, Sidahmed, Garrett, Wu, Rush, and
  Prakash]{singhal2021federated}
Karan Singhal, Hakim Sidahmed, Zachary Garrett, Shanshan Wu, Keith Rush, and
  Sushant Prakash.
\newblock Federated reconstruction: Partially local federated learning.
\newblock \emph{arXiv preprint arXiv:2102.03448}, 2021.

\bibitem[Tram{\`e}r and Boneh(2020)]{tramer2020differentially}
Florian Tram{\`e}r and Dan Boneh.
\newblock Differentially private learning needs better features (or much more
  data).
\newblock \emph{arXiv preprint arXiv:2011.11660}, 2020.

\bibitem[Vaswani et~al.(2017)Vaswani, Shazeer, Parmar, Uszkoreit, Jones, Gomez,
  Kaiser, and Polosukhin]{vaswani2017attention}
Ashish Vaswani, Noam Shazeer, Niki Parmar, Jakob Uszkoreit, Llion Jones,
  Aidan~N. Gomez, Lukasz Kaiser, and Illia Polosukhin.
\newblock Attention is all you need, 2017.

\bibitem[Wang et~al.(2021{\natexlab{a}})Wang, Agarwal, and
  Papailiopoulos]{wang2021pufferfish}
Hongyi Wang, Saurabh Agarwal, and Dimitris Papailiopoulos.
\newblock Pufferfish: Communication-efficient models at no extra cost.
\newblock \emph{arXiv preprint arXiv:2103.03936}, 2021{\natexlab{a}}.

\bibitem[Wang et~al.(2021{\natexlab{b}})Wang, Charles, Xu, Joshi, McMahan,
  Aguera~y Arcas, Al-Shedivat, Andrew, Avestimehr, Daly, et~al.]{wang2021field}
Jianyu Wang, Zachary Charles, Zheng Xu, Gauri Joshi, H~Brendan McMahan, Blaise
  Aguera~y Arcas, Maruan Al-Shedivat, Galen Andrew, Salman Avestimehr,
  Katharine Daly, et~al.
\newblock A field guide to federated optimization.
\newblock \emph{arXiv:2107.06917}, 2021{\natexlab{b}}.

\bibitem[Wang et~al.(2020)Wang, Li, Khabsa, Fang, and Ma]{wang2020linformer}
Sinong Wang, Belinda~Z. Li, Madian Khabsa, Han Fang, and Hao Ma.
\newblock Linformer: Self-attention with linear complexity, 2020.

\bibitem[Wieting and Kiela(2019)]{wietingrandom2019}
John Wieting and Douwe Kiela.
\newblock No training required: Exploring random encoders for sentence
  classification.
\newblock \emph{CoRR}, abs/1901.10444, 2019.
\newblock URL \url{http://arxiv.org/abs/1901.10444}.

\bibitem[Yang et~al.(2019)Yang, Liu, Chen, and Tong]{yang2019federated}
Qiang Yang, Yang Liu, Tianjian Chen, and Yongxin Tong.
\newblock Federated machine learning: Concept and applications.
\newblock \emph{ACM Transactions on Intelligent Systems and Technology (TIST)},
  10\penalty0 (2):\penalty0 1--19, 2019.

\bibitem[Zhang et~al.(2019)Zhang, Bengio, and Singer]{zhang2019layers}
Chiyuan Zhang, Samy Bengio, and Yoram Singer.
\newblock Are all layers created equal?, 2019.

\bibitem[Zhou et~al.(2020)Zhou, Wu, and Banerjee]{zhou2020bypassing}
Yingxue Zhou, Zhiwei~Steven Wu, and Arindam Banerjee.
\newblock Bypassing the ambient dimension: Private sgd with gradient subspace
  identification.
\newblock \emph{arXiv preprint arXiv:2007.03813}, 2020.

\end{thebibliography}

\newpage
\appendix

\section{Datasets}
\paragraph{CIFAR-10}
The CIFAR-10 dataset \citep{krizhevsky2009learning} consists of images with 3 channels of 32 x 32 pixels each.
Each pixel is represented by an int8.
This dataset contains 50,000 training examples, and 10,000 test examples.
The 10 possible labels are equally represented among the images (with 6,000 images per label).
While this dataset does not have a natural partition by client, we follow \citep{reddi2020adaptive}, and use the approach described in \citep{hsu2019measuring} to federate it.
We randomly partition the training data into 500 clients, each receiving 100 examples.
We apply Latent Dirichlet Allocation over the labels, and each client draws a multinomial distribution over the labels from a symmetric Dirichlet distribution with parameter $\alpha$.
This method results in an IID split when $\alpha\xrightarrow{}\infty$, while each client tends to draw a single label when $\alpha\xrightarrow{}0$.
We set $\alpha=1$ in our experiments.
We preprocess the images by randomly cropping each channel of the training images to shape $(24, 24)$, and randomly flipping them horizontally.
The test images are centrally cropped to the same dimensions.
We normalize the pixel values by centering them, and dividing them by their standard deviations.

\paragraph{EMNIST}
We use a federated version of the EMNIST dataset, where the characters are partitioned by their authors \citep{caldas2019leaf}. There are 62 possible characters, split among 3,400 clients.

\paragraph{Stack Overflow}
The Stack Overflow dataset contains the body text of questions and answers from the Stack Overflow website.
This data is available for 342,477 training clients with 135,818,730 examples, 38,758 validation clients with 16,491,230 examples, and 204,088 test clients with 16,586,035 examples.
It is naturally partitioned into clients by the post authors.
The train clients only have examples timestamped before 2018-01-01 UTC, and the test clients examples from after 2018-01-01 UTC.
The validation clients are held out from both train, and test sets.
Users with less than 100 training examples are filtered out.
We limit our vocabulary to the 10,000 most frequent words, and map the other words to an out-of-vocabulary bucket.
We restrict each client to the first 256 sentences.
We pad and truncate each sentence to ensure it has a length of 20 words.

\section{Models}
We use the CNN model described in \citep{mcmahan2017communication} to train EMNIST, adding a group normalization layer after the second convolution.
A full description of this model is given in table \ref{emnist-model}.
We freeze the first dense layer in our experiments with the corresponding partially trainable model.

\begin{table}
  \caption{EMNIST model.}
  \label{emnist-model}
  \centering
  \begin{tabular}{lll}
    \toprule
    Layer                       &       Output Shape              &   Parameters            \\
    \midrule
    Input                       &       $(28, 28, 1)$             &   $0$                   \\
    Conv2d                      &       $(28, 28, 32)$            &   $832$                 \\
    MaxPool2d                   &       $(14, 14, 32)$            &   $0$                   \\
    Conv2d                      &       $(14, 14, 64)$            &   $51264$               \\
    GroupNorm                   &       $(14, 14, 64)$            &   $128$                 \\
    MaxPool2d                   &       $(7, 7, 64)$              &   $0$                   \\
    Flatten                     &       $3136$                    &   $0$                   \\
    Dense                       &       $512$                     &   $1606144$             \\
    Dense                       &       $62$                      &   $31806$               \\
    \bottomrule
  \end{tabular}
\end{table}

We train ResNet-18 on CIFAR-10, where we replaced the batch normalizations by group normalizations, following the observation by \citep{hsieh2020noniid} that this change results in better behavior in non-idd settings.
To partially train this model, we freeze the convolutional layers of the residual blocks in increasing order.

We perform next word prediction on the Stack Overflow dataset using a three layer Transformer architecture \citep{vaswani2017attention}.
The token embeddings have a dimension set to 96, and the hidden dimension of the feed forward network (FFN) block to 2,048.
The multi-head attentions have 8 heads, each based on 12-dimensional key, query, value vectors. 
The activation used is ReLU, and the dropout rate is set to 0.1 during training.
The partially trainable models are derived by freezing the hidden layers of the FFN.

\section{Experiment Details}
We run our experiments using the Tensorflow Federated framework \citep{tff2019}, on GPUs.
We conduct 5 runs for each setting, with different random seeds, except in our DP-FTRL experiments, which we ran only once.
We present the average final accuracies, $\pm$ the standard deviations over these runs.
For each experiment, we use the SGD optimizer for the server updates.

We summarize more experimental details in tables \ref{experiment-details} and \ref{dp-ftrl-experiment-details}.

\begin{table}
  \caption{Experiment details.}
  \label{experiment-details}
  \centering
  \begin{tabular}{lllll}
    \toprule
    Task                &       Training rounds     &   Clients per round   &   Batch Size  &   Client Optimizer  \\
    \midrule
    EMNIST              &       $1500$              &   $20$                &   $16$        &   SGD               \\
    CIFAR-10            &       $1500$              &   $10$                &   $128$       &   SGDM              \\
    SO NWP              &       $5000$              &   $32$                &   $16$        &   Adam              \\
    \bottomrule
  \end{tabular}
\end{table}

\begin{table}
  \caption{DP-FTRL experiment details.}
  \label{dp-ftrl-experiment-details}
  \centering
  \begin{tabular}{llll}
    \toprule
    Task                &       Training rounds     &   Report Goal   &   Batch Size       \\
    \midrule
    SO NWP              &       $1600$              &   $100$                &   $16$                \\
    \bottomrule
  \end{tabular}
\end{table}

\subsection{Hyperparameters}
We present the hyperparameters used in our experiments.
\paragraph{Non DP experiments}
Table \ref{hparams-details} details the optimization parameters we use in all our experiments, except for the DP-FTRL ones.
The server momentum value used in CIFAR-10 is 0.9.

\begin{table}
  \caption{Optimizers.}
  \label{hparams-details}
  \centering
  \begin{tabular}{lllll}
    \toprule
    Task                &       Server Optimizer     &   Client Learning rate   &   Server learning rate  \\
    \midrule
    EMNIST              &       SGD              &   $0.05$                &   $0.5$        \\
    CIFAR-10            &       SGDM              &   $10^{-0.5}$                &   $10^{-1.0}$       \\
    SO NWP              &       Adam              &   $0.1$                &   $0.03$        \\
    \bottomrule
  \end{tabular}
\end{table}

\paragraph{DP-FTRL experiments}
The DP-FTRL experiments were run on Stack Overflow only.
We used the SGD optimizer on the clients, and SGD with a momentum value of 0.9 on the server.
We set the clipping norm to 0.3.
We performed a grid search on the learning rates for each noise multiplier value, and report the best accuracies.
The grid search was performed on the following values:

$$\eta_{client} \in \{10^{-1.5}, 10^{-1.0}, 10^{-0.5}\}$$
$$\eta_{server} \in \{10^{-1.5}, 10^{-1.0}, 10^{-0.5}, 10^{0.0}, 10^{0.25}\}$$

\subsection{Frozen/Trainable parameters}
We now detail the layers that were frozen in each set of experiments.

\paragraph{EMNIST}
The majority of the parameters in the model we use to do character recognition on EMNIST is contained in the dense layer following the convolutional blocks.
We freeze this layer when experimenting with a partially trained version of that model.

\paragraph{CIFAR-10}
We vary the percentage of trainable parameters in the Resnet-18 model we train on CIFAR-10 by increasing the number of frozen convolutional blocks, as shown in table \ref{resnet-18-frozen}.

\begin{table}
  \caption{Frozen ResNet-18 convolutional blocks.}
  \label{resnet-18-frozen}
  \centering
  \begin{tabular}{lllll}
    \toprule
    Trainable parameters (\%)                &       Frozen convolutional blocks       \\
    \midrule
    $100$              &       None                      \\
    $26.25$              &       $0$                      \\
    $8.07$              &       $0, 1$                      \\
    $3.47$            &       $0, 1, 2$                     \\
    $2.16$              &       $0, 1, 2, 3$                      \\
    \bottomrule
  \end{tabular}
\end{table}

\paragraph{Stack Overflow}
The Transformer architecture we use in our Stack Overflow Next Word Prediction experiments contains three encoder layers.
We vary the proportion of trainable parameters by freezing the first layer of the FFNs of these encoders, as detailed in table \ref{transfo-frozen}.

\begin{table}
  \caption{Partially frozen encoder layers in SO NWP.}
  \label{transfo-frozen}
  \centering
  \begin{tabular}{lllll}
    \toprule
    Trainable parameters (\%)                &       Partially frozen encoder blocks       \\
    \midrule
    $100$              &       None                      \\
    $91.3$              &       $2$                      \\
    $82.6$            &       $1, 2$                     \\
    $73.8$              &       $0, 1, 2$                      \\
    \bottomrule
  \end{tabular}
\end{table}

\subsection{Runtimes}
We report the average training time per round from our TFF simulation experiments.
We filtered out aberrant values by removing any value deviating by more than one standard deviation of the average in one run.

\end{document}